%% file: iclr2027_conference.tex
\documentclass{article} 
\usepackage{iclr2027_conference,times}

\input{math_commands.tex}

\usepackage{mathtools}
\usepackage{amssymb,amsthm}
\usepackage{url}
\usepackage{graphicx}
\usepackage{subcaption}
\usepackage{wrapfig}
\usepackage{booktabs}
\usepackage{multirow}
\usepackage{hyperref}
\usepackage{enumitem}
\usepackage{xspace}
\usepackage{cleveref}
\usepackage[section]{placeins}
\newsavebox{\legendbox}

\theoremstyle{plain}

\theoremstyle{definition}

\theoremstyle{remark}

\title{
Local Entropy Fluctuations for 
Black-Box 
LLM-Generated Text Detection
}

\author{Xinpeng Wang\footnotemark[1]\ ,\hspace{0.8em} Bruce Rushing\thanks{Equal contribution.}\ ,\hspace{0.8em} Zhizhen Chen,\hspace{0.8em} Zhe Zeng \\
University of Virginia
}

\newcommand{\methodPPR}{PPR\xspace}
\newcommand{\methodPPRminus}{DiffPPR\xspace}
\newcommand{\methodPPRoverE}{RelPPR\xspace}
\newcommand{\methodPPRminusFull}{Difference Peak Prominence Rate (\methodPPRminus)\xspace}
\newcommand{\methodPPRoverEFull}{Relative Peak Prominence Rate (\methodPPRoverE)\xspace}
\newcommand{\methodPPRS}{DiffPPR-S\xspace}

\iclrfinalcopy 
\begin{document}

\maketitle

\begin{abstract}
Zero-shot detection of machine-generated text is vital for mitigating the risks of large language models (LLMs) misuse. While the state-of-the-art dual-model detectors and resampling-based detectors outperform single-model statistical detectors on unprompted continuations, they suffer a catastrophic breakdown on instruction-conditioned text in the black-box setting, where the source model is unknown.
We analyze this failure and trace it to poor prompt conditioning for resampling-based methods and diminished cross-model discrepancy for dual-model detectors.
Motivated by recent studies that human writing maintains a steady, uniform information flow, whereas machine-generated text exhibits sharp, localized entropy fluctuations, we propose a new metric, the \methodPPRoverEFull to capture this structural contrast along the entropy trajectory and further leverage it as a signal for LLM-generated text detection.
Extensive evaluations across diverse model families and datasets demonstrate that \methodPPRoverE achieves state-of-the-art black-box detection performance and remains the most reliable detector under various practical situations. Our code is available at \url{https://github.com/brushing-git/entropy-llm-detector}.
\end{abstract}
\section{Introduction}

The rapid development of large language models (LLMs) capable of producing fluent, human-like text at scale has raised growing concerns about misuse~\citep{pmlr-v306-wang26r}, making reliable detection of machine-generated text an  urgent problem. We refer to the model that generate candidate text as the \emph{source model}. 
Detection under the \emph{black-box setting} assumes access only to the candidate text itself, with no knowledge of the source model's identity or parameters. This setting requires neither labeled training data nor access to the source model, making it a highly practical yet challenging paradigm. Detectors instead rely on one or more surrogate models, which we refer to as \emph{evaluator models}, to determine whether the candidate text is LLM-generated.

Existing zero-shot detectors broadly fall into three families: (i) \emph{single-model statistical} criteria, which score a candidate text using only a single evaluator model's statistics, such as log-likelihood, log-rank, and entropy~\citep{log-likelihood,ippolito2020automatic,logrank}; (ii) \emph{resampling-based} detectors, which regenerate the text under an evaluator model to  compare the candidate against this regeneration or normalize specific statistics~\citep{detectgpt,dnagpt,lastde}; and (iii) \emph{dual-model detectors}, which rely on the discrepancy between two evaluator models' probability estimates for the candidate text, typically measured via statistics such as log-likelihood or cross-entropy~\citep{fastdetectgpt,binoculars,dna-detectllm}. The latter two families currently define the state of the art, substantially outperforming single-model statistics.

\begin{figure}[t]
\begin{subfigure}[t]{0.48\textwidth}
        \centering
        \includegraphics[width=\textwidth]{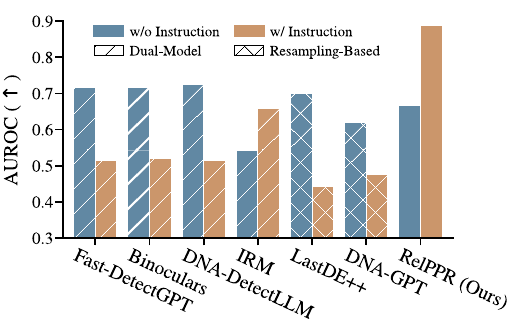}
        \caption{AUROC under unprompted and instruction-conditioned generation.}
        \vspace{-2pt}
        \label{fig:project_motivation}
    \end{subfigure}
    \hfill
    \begin{subfigure}[t]{0.48\textwidth}
        \centering
        \includegraphics[width=\textwidth]{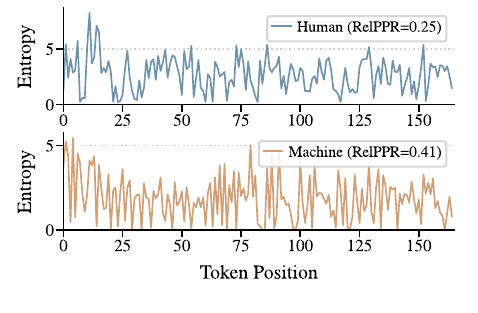}
        \caption{The entropy trajectory of human and machine text in response to the same question from ELI5 dataset.}
        \vspace{-2pt}
        \label{fig:method_motivation}
    \end{subfigure}
    
    \caption{(a)  \methodPPRoverE is robust under instruction-conditioning where existing detectors degrade; (b) Machine-generated text exhibits stronger entropy fluctuations than human text.}
    \vspace{-1.0em}
\end{figure}

We find, however, that this advantage holds mainly for \emph{unprompted} continuations of a human prefix, i.e., text generated without user prompt or instruction. Once text is generated under the instruction-conditioned regime, dual-model and resampling-based detectors alike suffer a serious performance degradation as shown in \Cref{fig:project_motivation}. As our first contribution, we analyze this failure mode systematically for both detector families. 1) For resampling-based methods, since the user prompt is invisible at evaluation time, they cannot resample text by conditioning on the original prompt as the source model did. This \emph{prompt conditioning gap} drives the resampled text to drift away from the candidate text. 2) For dual-model detectors, the cross-model discrepancy they rely on collapses under instruction conditioning due to a systematic shrinkage in cross-entropy, which is driven  by lower predictive uncertainty (entropy) or by heightened cross-model consensus through reduced KL divergence.
In summary, the prompt conditioning gap and the diminished cross-model discrepancy undermine predictive power in resampling-based and dual-model detectors, respectively.

Since all existing state-of-the-art (SOTA) detectors fail under instruction conditioning, our goal is to
find a new detection signal that is effective under such setting. According to the \emph{Uniform Information Density} (UID) principle in psycholinguistics~\citep{jaeger2010redundancy,meister2021revisiting}, humans naturally regulate their writing to maintain a steady rate of information transmission, producing a smooth information flow with restrained local fluctuations. In contrast, LLMs concentrate uncertainty on semantic branch points and key decision junctures, a pattern further accentuated by instruction-tuning~\citep{wang2025beyond,wei2026entropy,bigelow2025forking}. Therefore, we hypothesize that machine text exhibits higher-amplitude local fluctuations in its entropy trajectory than human text. 

Motivated by the above observations and hypothesis, we propose the Peak Prominence Rate (\methodPPR), defined as the average prominence of local peaks along the entropy trajectory, to quantify trajectory fluctuations for detection. However, a text with higher sequence-level mean entropy can attain a larger raw
\methodPPR even if its fluctuations are mild relative to its entropy level. To counteract this bias, we normalize \methodPPR by the sequence-level mean entropy, yielding our final criterion, \textbf{\methodPPRoverE}. As illustrated in \Cref{fig:method_motivation}, the entropy trajectory of machine text exhibits sharper fluctuations than that of human text relative to their respective entropy levels, yielding a higher \methodPPRoverE. Notably, \methodPPRoverE is entirely parameter-free and requires only a single evaluator model during inference.

We evaluate \methodPPRoverE across seven datasets and five source models under instruction-conditioning. Beyond achieving the state-of-the-art detection performance under the black-box setting, \methodPPRoverE demonstrates strong practical deployment value: it achieves superior recall under strict false-positive constraints, sustains reliability across varying sequence lengths and decoding configurations, withstands adversarial paraphrasing, and generalizes across diverse evaluator model scales and families.

Our contributions are threefold:
    (1) We show that dual-model and resampling-based detectors degrade catastrophically on instruction-conditioned text, and we trace this degradation to concrete mechanisms: diminished cross-model discrepancy for dual-model detectors and the prompt conditioning gap for resampling-based detectors.
    (2) We propose \methodPPRoverE, a single-model, parameter-free detection criterion motivated by the observation that machine-generated text exhibits higher-amplitude fluctuations in entropy trajectory than human text.
    (3) We conduct extensive black-box evaluations across seven datasets and five source models, showing that \methodPPRoverE achieves the state-of-the-art performance and remains the most reliable detector across diverse deployment conditions.

\section{Related Work}

\textbf{Zero-Shot Detectors} Unlike supervised classifiers that struggle 
with out-of-distribution generalization~\citep{pu2023zero, 
doughman2025exploring}, zero-shot detectors require no model cooperation 
or labeled data, and fall into three paradigms. \textbf{Single-model 
statistical criteria} score text by sequence-level statistics under one 
evaluator model, from simple metrics such as likelihood~\citep{log-likelihood}, 
LogRank~\citep{logrank}, and entropy~\citep{ippolito2020automatic}, to 
refinements using metric ratios (LRR~\citep{LRR}),  likelihood trajectroy 
dynamics (LastDE~\citep{lastde}), or grammatical edit features 
(GECScore~\citep{gecscore}). \textbf{Resampling-based detectors} instead 
evaluate fidelity by regenerating content from the observed context.
DNA-GPT~\citep{dnagpt} directly compares the regenerated text with the
candidate via $n$-gram overlap, while LastDE++~\citep{lastde} uses the
regenerated samples only to normalize its Lastde score.
\textbf{Dual-model detectors} construct a dynamic 
baseline from a second model rather than a fixed statistic: DetectGPT~\citep{detectgpt} 
and Fast-DetectGPT~\citep{fastdetectgpt} obtain this baseline via 
perturbation and token substitution respectively, Binoculars~\citep{binoculars} 
and DNA-DetectLLM~\citep{dna-detectllm} normalize surprisal against it, and 
IRM~\citep{IRM} instead baselines against a base (unaligned) model.
While dual-model and resampling-based detectors degrade under instruction conditioning, our proposed \methodPPRoverE performs exceptionally well in this challenging regime by capturing the fine-grained local fluctuation structure of the entropy trajectory.

\textbf{Entropy Dynamics and Uniform Information Density}
\emph{Uniform Information Density} hypothesis from psycholinguistics \citep{jaeger2010redundancy,NIPS2006_c6a01432,meister2021revisiting} posits that humans intuitively spread out the information in communication, avoiding sudden bursts of hard-to-predict content so that listeners can follow along smoothly.
In contrast, LLM generations concentrate uncertainty on a few pivotal ``forking'' tokens. \citet{bigelow2025forking} locate such tokens by resampling from each prefix and show that a single token can redirect the whole generation; \citet{wang2025beyond} find that these high-entropy tokens are a small minority in chain-of-thought and that restricting RLVR updates to them suffices; \citet{wei2026entropy} branch the decoding tree only at high-entropy positions, treating them as natural points for exploration. These works study entropy dynamics to understand or improve generation. In contrast, we use entropy dynamics for black-box detection: the entropy trajectory of machine-generated text fluctuates more sharply than that of human writing. To our knowledge, this signal has not been explored for detection.

\section{What Breaks Current Detectors under User Instructions}\label{sec:dual-model-fail}\label{sec:baseline-fail}

Empowered by post-training alignment, instruction-tuned LLMs produce highly fluent and task-conforming text under instructions, making detection harder than in unprompted continuation. We thus evaluate whether current zero-shot detectors remain effective in this demanding regime. 
Specifically, we contrast two generation settings on the \emph{same} human texts: (i) \emph{w/o instruction}, where the model performs unprompted continuation from a bare prefix, and (ii) \emph{w/ instruction}, where the prefix is wrapped into task-specific prompt templates as shown in \Cref{tab:appendix-prompt-templates}. 
We set the source model as OLMo-3.1-32B-Instruct and evaluate detectors under both settings. 
As shown in Figure~\ref{fig:project_motivation}, current detectors degrade severely under instruction conditioning.
Below, we uncover the underlying mechanisms driving this breakdown across both dual-model and resampling-based methods.

\subsection{Preliminaries and Notations}\label{sec:notations}

Let $\mathcal{V}$ denote the token vocabulary. Given a candidate text sequence $s = (x_1, \dots, x_L) \in \mathcal{V}^L$ of length $L$, we write the prefix preceding position $i$ as $x_{<i} = (x_1, \dots, x_{i-1})$. 
Given model $p$, we denote $p(\cdot \mid x_{<i})$ as the conditional distribution over the next token $x_i\in \mathcal{V}$ given the prefix $x_{<i}$.
We denote the sequence-level surprisal $I_{p}(s)$, entropy $H(p;s)$, and cross-entropy $H(p_1, p_2; s)$ for evaluator models $p_1$ and $p_2$ as follows, respectively:
\begin{gather*}
I_{p}(s) = -\frac{1}{L}\sum_{i=1}^L \log p(x_i \mid x_{<i}),
\quad
H(p;s) = -\frac{1}{L}\sum_{i=1}^L \sum_{v \in \mathcal{V}} p(v\mid x_{<i})\log p(v\mid x_{<i}), \\
H(p_1, p_2; s) = -\frac{1}{L}\sum_{i=1}^L \sum_{v \in \mathcal{V}} p_1(v\mid x_{<i})\log p_2(v\mid x_{<i}).
\end{gather*}

\subsection{Dual-Model Detectors}\label{sec:dual-model-detectors}

\textbf{Fast-DetectGPT}
\label{para:fastdetectgpt}\label{sec:fast-detectgpt}
Fast-DetectGPT~\citep{fastdetectgpt} refines the probability curvature idea of
DetectGPT~\citep{detectgpt}: a model's own generations tend to be more probable
than nearby alternatives (e.g., slight rewrites), whereas human text shows no
such tendency.
Instead of DetectGPT's costly perturb-and-rescore procedure, Fast-DetectGPT operates at the token level: at each position, the observed token $x_i$ is compared against alternatives sampled from $p_1(\cdot\mid x_{<i})$ and scored by $p_2$. Because these alternatives are sampled independently given the fixed passage, their expected score admit an analytic solution,
\begin{equation}\label{eq:fastdetectgpt-score}
  \mathrm{Score}_{\mathrm{F}}(s) = \frac{N(s)}{\sqrt{\sum_{i=1}^L \tilde\sigma_i^2}}, \qquad N(s) = L\cdot \big(H(p_1, p_2; s) - I_{p_2}(s)\big),
\end{equation}
where $H(p_1, p_2; s)$ is the sequence-level cross-entropy between $p_1$ and $p_2$, and $I_{p_2}(s)$ is the surprisal under $p_2$, detailed in \Cref{sec:appendix-fast-detectgpt}.
We show in \Cref{tab:fast_detectgpt_numerator_ablation} that dropping $\sqrt{\sum_{i=1}^L \tilde\sigma_i^2}$ in $\mathrm{Score}_{\mathrm{F}}(s)$ has negligible impact on
detection performance, and thus we focus on $N(s)$ in the subsequent analysis.

\textbf{Binoculars and DNA-DetectLLM}
\label{para:binoculars}
Instead of taking the difference, Binoculars~\citep{binoculars} and DNA-DetectLLM~\citep{dna-detectllm} score text via a ratio between surprisal and cross-entropy, denoted by $\mathrm{Score}_{\mathrm{B}}(s)$ and $\mathrm{Score}_{\mathrm{D}}(s)$ respectively, with details in \Cref{sec:appendix-binoculars-dna-detectllm}:
\begin{equation}\label{eq:binoculars-dna-detectllm}
\mathrm{Score}_{\mathrm{B}}(s) = -\frac{I_{p_2}(s)}{H(p_1, p_2; s)},  \qquad \mathrm{Score}_{\mathrm{D}}(s) = -\frac{I_{p_2}(s)+I_{p_2}(\tilde{s} \mid s )}{2\cdot H(p_1, p_2; s)}.
\end{equation}
In $\mathrm{Score}_{\mathrm{D}}(s)$, $\tilde{s}_t = \arg\max_{v \in \mathcal{V}} p_2(v \mid s_{<t})$ is the greedy repair of $s$, and $I_{p_2}(\tilde{s} \mid s)$ measures the minimum attainable surprisal along the prefixes of $s$. 
Under the premise of DNA-DetectLLM, machine-generated text is hypothesized to adhere closely to the greedy decoding trajectory of LLM, so that $I_{p_2}(\tilde{s} \mid s)$ approximates $I_{p_2}(s)$ for machine text while deviating substantially for human text. 



\begin{figure}[t]
\centering
\includegraphics[width=\textwidth]{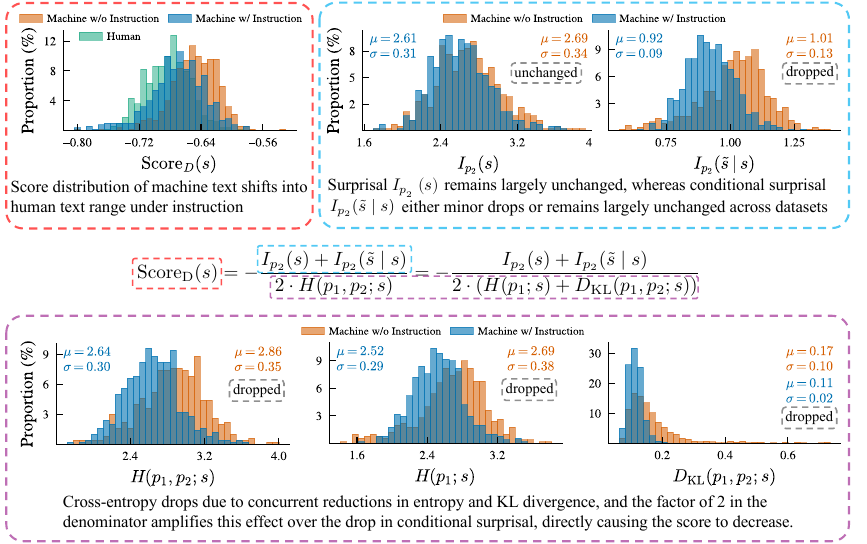}
\caption{Analysis of the DNA-DetectLLM score $\mathrm{Score}_{\mathrm{D}}(s)$ on XSum, with OLMo-3.1-32B-Instruct as the source model. $\mu$ and $\sigma$ denote the mean and standard deviation.
Under instruction conditioning, machine text scores shift lower, converging toward the human score distribution.  
}
\vspace{-1em}
\label{fig:statistic-analysis}
\end{figure}

\textbf{Unified Information-Theoretic Perspective}
All three methods operate on the premise that machine-generated text yields higher scores than human text. 
However, taking XSum as a representative case study, the detection statistics of machine-generated text drop noticeably across detectors under instruction conditioning, and their distributions shi ft closer to that of human text, as illustrated for DNA-DetectLLM in \Cref{fig:statistic-analysis} and for the Fast-DetectGPT numerator $N(s)$ alongside Binoculars in \Cref{fig:dual-model-baselines-xsum}. This distributional shift directly drives their performance degradation. We further analyze this phenomenon from an information-theoretic perspective.

Comparing \Cref{eq:fastdetectgpt-score,eq:binoculars-dna-detectllm}, we find that these three methods are governed by the \emph{two information-theoretic quantities}: the sequence-level surprisal $I_{p_2}(s)$ and the sequence-level cross-entropy $H(p_1, p_2; s)$.
As structurally decomposed in \Cref{fig:statistic-analysis}, the cross-entropy $H(p_1, p_2; s)$ shifts toward lower values under instruction conditioning, while the surprisal $I_{p_2}(s)$ remains largely unchanged.
Consequently, this asymmetric shift compresses the Fast-DetectGPT numerator $N(s)$ and the Binoculars score $\mathrm{Score}_{\mathrm{B}}(s)$ of machine text under instruction conditioning, as illustrated in \Cref{fig:dual-model-baselines-xsum}.
This phenomenon is consistently observed across other datasets, as detailed in \Cref{sec:appendix-fdgpt-components-other-datasets}.
For DNA-DetectLLM, the mean of the $I_{p_2}(\tilde{s} \mid s)$ distribution exhibits little variation across most datasets. Even on datasets where it shifts toward lower values, such as XSum in \Cref{fig:statistic-analysis}, its decrease does not substantially outpace the shrinkage of the cross-entropy denominator, ultimately driving the distribution of $\mathrm{Score}_{\mathrm{D}}(s)$ toward lower values as well. 

To understand the shrinkage of cross-entropy at a finer granularity, we decompose the sequence-level cross-entropy into model entropy and divergence:
\begin{equation*}
  H(p_1, p_2; s) = H(p_1; s) + D_{\mathrm{KL}}(p_1, p_2; s),
\end{equation*}
where $D_{\mathrm{KL}}(p_1, p_2; s)$ denotes the sequence-level KL divergence, with derivation provided in \Cref{sec:appendix-cross-entropy-decomposition}.
As shown in \Cref{fig:statistic-analysis}, the cross-entropy shrinkage on XSum stems from concurrent drops in both KL divergence and model entropy.
Across other datasets, this shrinkage is driven either predominantly by a drop in divergence $D_{\mathrm{KL}}(p_1, p_2; s)$ on CNN/DailyMail and WikiText, or primarily by a reduction in model entropy $H(p_1; s)$ on SQuAD and ELI5.

In summary, dual-model detectors degrade directly due to cross-entropy shrinkage, showing that cross-model discrepancy fails under instruction conditioning. Specifically, this breakdown is driven either by lower predictive uncertainty (lower entropy $H(p_1; s)$), or by heightened cross-model consensus through reduced KL divergence $D_{\mathrm{KL}}(p_1, p_2; s)$.

\subsection{Resampling-Based Detectors}\label{sec:resampling-fail}
During text generation, the source model produces candidate text sequence $s$ from an initial human prefix $x_{\mathrm{pre}}$.
The source model generates tokens from $p_{\mathrm{src}}(\cdot \mid x_{<i}, x_{\mathrm{pre}})$ without instruction, whereas under instruction conditioning it samples from $p_{\mathrm{src}}(\cdot \mid x_{<i}, x_{\mathrm{pre}}, c)$ guided by an explicit instruction prompt $c$.
At test time, resampling-based detectors evaluate candidate text $s$ by generating alternative continuations $\tilde{x}_i \sim p(\cdot \mid x_{<i})$ using evaluator model $p$, and either compare them directly against $s$, such as via $n$-gram overlap in DNA-GPT~\citep{dnagpt}, or use them to calibrate a statistic on $s$, as in LastDE++~\citep{lastde}.
Because neither the prefix $x_{\mathrm{pre}}$ nor the instruction $c$ is accessible to the evaluator, the unconditioned resampling distribution $p(\cdot \mid x_{<i})$ deviates far more drastically from the true generative distribution $p_{\mathrm{src}}(\cdot \mid x_{<i}, x_{\mathrm{pre}}, c)$ under instruction conditioning than from $p_{\mathrm{src}}(\cdot \mid x_{<i}, x_{\mathrm{pre}})$ in unprompted continuation.
This widened prompt conditioning gap causes the resampled continuations to drift heavily away from the instruction-guided trajectory of $s$, depriving resampling-based detectors of their discriminative signal.

\section{Our Method: Relative Peak Prominence Rate}
\label{sec:pre}
To address the challenges posed by instruction conditioning, we draw on the difference in how humans and machines produce text and propose to detect machine-generated text by the difference of fluctuation in its entropy trajectory. We capture this fluctuation signal using \emph{topological prominence}, which measures how sharply each local entropy peak stands out above its neighboring valleys. 

\subsection{Motivation: Human Text Flows, Machine Text Spikes}

\label{sec:method-motivation}
\textbf{Human Writing: Smooth Information Flow} As formalized in the \emph{Uniform Information Density} (UID) principle~\citep{jaeger2010redundancy,meister2021revisiting}, humans naturally regulate their writing to keep the rate of information transmission relatively steady (e.g., by cushioning complex concepts with preparatory context to facilitate reader comprehension). This communicative buffering yields a smooth information flow that inherently avoids sharp, localized entropy spikes.
Extending this principle to generation dynamics, we posit that the resulting entropy trajectory of human text exhibits smoother transitions with restrained local fluctuations.

\textbf{Machine Text: High-Amplitude Local Fluctuations} Autoregressive generation in LLMs, in
contrast, is not globally planned for smooth communication. Prior work shows uncertainty
varies across positions: certain high-entropy tokens, termed \emph{forking tokens}, act as decision
points steering the continuation~\citep{wang2025beyond}. They often coincide with logical
connectives (e.g., ``however'', ``thus'') and question words~\citep{wei2026entropy},
and a single such token can redirect the whole generation~\citep{bigelow2025forking}.
We posit that the alternation between forking tokens and predictable stretches yields an entropy
trajectory with higher-amplitude local fluctuations than human text.

\textbf{Instruction-Conditioned Text: Even Stronger Fluctuations}
\citet{wang2025beyond} show that reinforcement learning with verifiable rewards (RLVR) does not reshape a
model's token-level entropy uniformly: tokens with higher initial entropy show larger entropy increases after RLVR,
while the entropy of low-entropy tokens stays close to its pre-RLVR level.
Motivated by this, and since RLVR is run on instruction prompts, we conjecture that instruction conditioning accentuates peaks at forking tokens, yielding larger fluctuations than unprompted continuation.

Overall, based on the distinct generation mechanisms of humans and machines, we conjecture that machine text generated with instructions exhibits larger fluctuations in its entropy trajectory than human text, and we therefore formulate our detection criterion by  quantifying these fluctuations.

\subsection{Peak Prominence Rate of Entropy Trajectory}\label{sec:pre-formal}
To capture the degree of local fluctuation as a detection signal, we adopt the concept of \emph{topological prominence} from computational geometry and signal analysis~\citep{perea2015sliding}. Given a text sequence $s$ and a evaluator model $M$, we compute the token-level entropy trajectory $\mathbf{H}_{M}(s) = (h_1, \dots, h_L) \in \mathbb{R}^L$, where $h_i = H\big(p(\cdot \mid x_{<i})\big)$ denotes the entropy at step $i$. A position $t \in \{2, \dots, L-1\}$ is a peak if $h_t > h_{t-1}$ and $h_t > h_{t+1}$; we denote the set of all such peaks by $\mathcal{T}(s)$. 

For each peak $t \in \mathcal{T}(s)$, its prominence $\mathrm{Prom}(t)$ is defined as the vertical distance between the peak and the highest contour line that connects it to an even higher peak: 
\begin{equation*}
\mathrm{Prom}(t) = h_t - \max\Big(\min_{i \in [l_t, t]} h_i,\, \min_{j \in [t, r_t]} h_j\Big),
\end{equation*}
where $l_t$ and $r_t$ are the nearest preceding and succeeding positions along the trajectory whose entropy exceeds $h_t$ (defaulting to the sequence boundaries if no such position exists). Thus, $\mathrm{Prom}(t)$ measures the elevation of peak $t$ above its local valley.

We define the sequence-level \emph{Peak Prominence Rate} (\methodPPR) as the cumulative prominence of all peaks normalized by the sequence length:
\begin{equation*}
\mathrm{\methodPPR}(s) = \frac{1}{L}\sum_{t \in \mathcal{T}(s)} \mathrm{Prom}(t).
\end{equation*}

As human text generally exhibits higher entropy than machine-generated text under the evaluator model~\citep{logrank}, a high-entropy human text can attain a large raw \methodPPR, even when its fluctuations are mild relative to its own entropy level.
Therefore, we introduce two entropy-adjusted variants: (i) the \methodPPRoverEFull  score, which normalizes \methodPPR by the sequence-level mean entropy $H(p;s)$ to isolate fluctuation relative to the sample's own entropy level; (ii) the \methodPPRminusFull score, which treats \methodPPR as an incremental signal to $H(p;s)$. The two criteria are defined as:
\begin{equation*}\label{eq:ppr-combined}
\mathrm{Score}_{\mathrm{\methodPPRoverE}}(s) = \frac{\mathrm{\methodPPR}(s)}{H(p;s)},  \quad \mathrm{Score}_{\mathrm{\methodPPRminus}}(s)= \mathrm{\methodPPR}(s) - H(p;s).
\end{equation*}


\begin{figure}[t]
\centering
\includegraphics[width=2.1in]{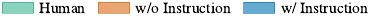}\\[-2pt]
\includegraphics[height=0.95in]{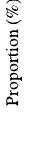}%
\begin{subfigure}[t]{1.31in}
  \centering
  \includegraphics[width=1.31in]{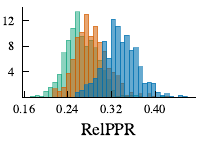}
  \captionsetup{margin={5pt,-5.5pt},skip=2pt}
  \caption{XSum}
\end{subfigure}%
\hfill%
\begin{subfigure}[t]{1.31in}
  \centering
  \includegraphics[width=1.31in]{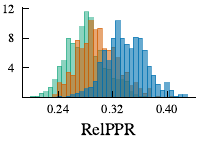}
  \captionsetup{margin={5pt,-5.5pt},skip=2pt}
  \caption{CNN/DailyMail}
\end{subfigure}%
\hfill%
\begin{subfigure}[t]{1.31in}
  \centering
  \includegraphics[width=1.31in]{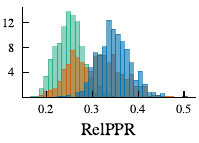}
  \captionsetup{margin={5pt,-5.5pt},skip=2pt}
  \caption{ELI5}
\end{subfigure}%
\hfill%
\begin{subfigure}[t]{1.31in}
  \centering
  \includegraphics[width=1.31in]{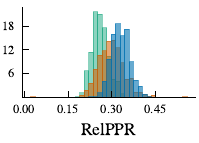}
  \captionsetup{margin={5pt,-5.5pt},skip=2pt}
  \caption{WritingPrompts}
\end{subfigure}
\caption{%
  Distributions of \methodPPRoverE across four datasets for human text alongside machine-generated text from OLMo-3.1-32B-Instruct. \methodPPRoverE consistently separates human writing and machine text generated under instruction conditioning. Results on additional datasets are provided in \Cref{fig:ppr-appendix}.
}
\vspace{-1em}
\label{fig:ppr-main}
\end{figure}

For both criteria, a higher \methodPPRoverE or \methodPPRminus score indicates a higher likelihood that the candidate text is machine-generated. 
In \Cref{fig:ppr-main}, we examine the empirical distributions of \methodPPRoverE across diverse benchmark datasets, holding the human-written prefix strictly constant between the two machine conditions.
Across all datasets, the resulting distributions systematically adhere to a consistent ordering: text generated under instruction conditioning attains the highest \methodPPRoverE values, followed by machine text without instruction, whereas human text consistently exhibits the lowest values.
This empirical ordering directly validates our three motivations outlined in \Cref{sec:method-motivation}, confirming that human text maintains a smooth information flow, machine text introduces sharp local variations at semantic branch points, and instruction conditioning further amplifies these localized fluctuations, accentuating rather than eroding the separation between human and machine distributions.

Compared with dual-model and resampling-based detectors, our criteria require only a
single forward pass and no sampling, making them computationally efficient, and, as shown in the
next section, they also achieve superior detection performance.

\section{Experiment}
In this section, we evaluate \methodPPRoverE against twelve zero-shot baselines across seven datasets 
and six instruction-tuned source models, and further examine its robustness to decision
thresholds, decoding configurations, generation length, paraphrasing attacks, and the size and family of evaluator model.
The  details regarding experimental setups are provided in \Cref{sec:appendix-implementation-details}.
\subsection{Experimental Setup}\label{sec: exp-setup}

\textbf{Benchmark Datasets}
We evaluate on seven benchmark datasets spanning diverse domains: news articles (XSum~\citep{xsum} and CNN/DailyMail~\citep{cnn-dailymail}), encyclopedic contexts (WikiText-103~\citep{wikitext} and SQuAD~\citep{squad}), creative story writing (WritingPrompts~\citep{writingprompts}), biomedical research question answering (PubMedQA~\citep{pubmedqa}) and general-topic question answering (ELI5~\citep{eli5}).

\textbf{Human Prefix and Prompt Templates} For datasets without a standalone question or opening sentence (XSum, CNN/DailyMail, WikiText, and SQuAD), following the Fast-DetectGPT protocol, the human prefix is formed by extracting the first $30$ tokens from the source document, while the remaining tokens serve as the human completion. For WritingPrompts, ELI5, and PubMedQA, the human prefix is the clean opening sentence or question itself, and the corresponding human continuation is the original human-written story or answer. 
Each dataset is equipped with a tailored instruction prompt template (detailed in \Cref{tab:appendix-prompt-templates}). Crucially, adhering to realistic zero-shot detection scenarios, the instruction prompt is strictly \textbf{invisible} to detectors during evaluation.

\textbf{Baselines and Evaluation Metric}
We evaluate our proposed method against $14$ competitive zero-shot baselines:
(1) \emph{Single-model statistical criteria:} Likelihood~\citep{log-likelihood}, Entropy~\citep{ippolito2020automatic}, LogRank~\citep{logrank}, LastDE~\citep{lastde}, LRR~\citep{LRR} and GECScore~\citep{gecscore};
(2) \emph{dual-model detectors:} Fast-DetectGPT~\citep{fastdetectgpt}, Binoculars~\citep{binoculars}, DNA-DetectLLM~\citep{dna-detectllm}, and Implicit Reward Model (IRM)~\citep{IRM};
(3) \emph{Resampling-based detectors:} DNA-GPT~\citep{dnagpt}  and LastDE++~\citep{lastde}.
Detection performance is measured by the Area Under the Receiver Operating Characteristic Curve (AUROC) across all human-machine pairs.
\begin{table}[t]
\centering
\footnotesize 
\renewcommand{\arraystretch}{0.9}
\caption{Detection performance (AUROC) by dataset (averaged across five source models) under black-box setting. The best and second-best results are shown in \textbf{bold} and \underline{underlined}, respectively.}
\label{tab:main_results_by_dataset}
\resizebox{0.95\linewidth}{!}{%
\begin{tabular}{@{} l ccccccc c @{}}
\toprule
\textbf{Methods} & \textbf{XSum} & \textbf{WritingP} & \textbf{SQuAD} & \textbf{WikiText} & \textbf{CNN/DM} & \textbf{ELI5} & \textbf{PubMedQA} & \textbf{Avg} \\
\midrule
Entropy & 0.822 & 0.812 & 0.813 & 0.788 & 0.817 & 0.929 & \textbf{0.703} & 0.812 \\
Likelihood & 0.787 & 0.801 & 0.787 & 0.752 & 0.775 & 0.902 & 0.600 & 0.772 \\
LogRank & 0.786 & 0.764 & 0.782 & 0.767 & 0.782 & 0.894 & 0.595 & 0.767 \\
LRR & 0.739 & 0.613 & 0.731 & 0.781 & 0.756 & 0.831 & 0.550 & 0.714 \\
GECScore & 0.712 & 0.673 & 0.684 & 0.648 & 0.688 & 0.839 & 0.558 & 0.686 \\
LastDE & 0.755 & 0.767 & 0.783 & 0.785 & 0.767 & 0.890 & 0.590 & 0.762 \\
LastDE++ & 0.406 & 0.545 & 0.432 & 0.317 & 0.433 & 0.673 & 0.444 & 0.464 \\
DNA-GPT & 0.401 & 0.509 & 0.372 & 0.441 & 0.335 & 0.568 & 0.486 & 0.445 \\
Fast-DetectGPT & 0.649 & 0.718 & 0.580 & 0.598 & 0.592 & 0.531 & 0.449 & 0.588 \\
Binoculars & 0.663 & 0.712 & 0.598 & 0.612 & 0.602 & 0.551 & 0.452 & 0.599 \\
DNA-DetectLLM & 0.704 & 0.731 & 0.590 & 0.646 & 0.614 & 0.420 & 0.451 & 0.594 \\
IRM & 0.867 & 0.807 & 0.707 & \underline{0.887} & 0.818 & 0.662 & 0.581 & 0.761 \\
\midrule
\methodPPR (Ours) & 0.715 & 0.787 & 0.555 & 0.610 & 0.638 & 0.461 & 0.412 & 0.597 \\
\methodPPRminus (Ours) & \underline{0.884} & \underline{0.909} & \underline{0.865} & 0.850 & \underline{0.869} & \underline{0.958} & \underline{0.670} & \underline{0.858} \\
\textbf{\methodPPRoverE} (Ours) & \textbf{0.921} & \textbf{0.961} & \textbf{0.888} & \textbf{0.913} & \textbf{0.879} & \textbf{0.974} & 0.563 & \textbf{0.871} \\
\bottomrule
\end{tabular}}
\vspace{-1em}
\end{table}

\textbf{Models}
To ensure robust evaluation across diverse model capacities and structural paradigms, we evaluate on five instruction-tuned LLMs spanning both dense and sparse Mixture-of-Experts (MoE) architectures, with parameter scales ranging from 7B to 119B.
We use the small Qwen3.5-2B~\citep{qwen3.5} as the evaluator in our main
experiments, simulating realistic black-box deployment scenarios where the
target commercial model substantially exceeds the detector in capacity.
We also evaluate evaluator models of different sizes and families in
\Cref{tab:evaluator_size_ablation}. Full model identities and evaluator configurations are detailed in \Cref{sec:appendix-detection-regimes}.

\subsection{Main Results}\label{sec:main_results}

We evaluate detectors under the black-box setting, where the detector has no access to the source model, and summarize their detection performance across datasets in \Cref{tab:main_results_by_dataset} and across source models in \Cref{tab:main_results_by_model}. The results demonstrate that our proposed \methodPPRoverE achieves state-of-the-art performances across diverse source model architectures and domains.

\begin{wraptable}[17]{r}{0.5\textwidth}
\vspace{-13pt}
\centering
\setlength{\tabcolsep}{2pt} 
\setlength{\belowcaptionskip}{2pt}
\caption{AUROC across source models, averaged over seven datasets under the black-box setting. The best and second-best results are shown in \textbf{bold} and \underline{underlined}, respectively.}
\label{tab:main_results_by_model}
\vspace{-6pt}
\resizebox{\linewidth}{!}{%
\begin{tabular}{lcccccc}
\toprule
\multirow{2}{*}{} & \textbf{OLMo3}          & \textbf{OLMo3.1}        & \textbf{Gemma4}         & \textbf{Mistral}         & \textbf{Qwen3.5}         & \multirow{2}{*}{\textbf{Avg}} \\
                          & \textbf{7B}              & \textbf{32B}             & \textbf{31B}             & \textbf{119B}            & \textbf{35B}             &                      \\ \midrule
Entropy                   & 0.780           & 0.837           & 0.811           & 0.841           & 0.790           & 0.812                \\
Likelihood                & 0.693           & 0.794           & 0.803           & 0.848           & 0.723           & 0.772                \\
LogRank                   & 0.699           & 0.789           & 0.786           & 0.838           & 0.723           & 0.767                \\
LRR                       & 0.687           & 0.740           & 0.692           & 0.763           & 0.689           & 0.714                \\
GECScore                  & 0.646           & 0.679           & 0.743           & 0.631           & 0.731           & 0.686                \\
LastDE                    & 0.704           & 0.781           & 0.764           & 0.817           & 0.745           & 0.762                \\
LastDE++                  & 0.406           & 0.442           & 0.476           & 0.558           & 0.440           & 0.464                \\
DNA-GPT                   & 0.462           & 0.475           & 0.430           & 0.454           & 0.401           & 0.445                \\
Fast-DetectGPT            & 0.420           & 0.515           & 0.693           & 0.712           & 0.601           & 0.588                \\
Binoculars                & 0.436           & 0.521           & 0.699           & 0.718           & 0.618           & 0.599                \\
DNA-DetectLLM             & 0.436           & 0.515           & 0.698           & 0.702           & 0.617           & 0.594                \\
IRM                       & 0.668           & 0.658           & \underline{0.887} & 0.698           & \textbf{0.895}  & 0.761                \\ \midrule
\methodPPR (Ours)                       & 0.586           & 0.575           & 0.638           & 0.568           & 0.616           & 0.597                \\
\methodPPRminus (Ours)                     & \underline{0.824} & \underline{0.877} & 0.874           & \underline{0.875} & 0.839           & \underline{0.858}      \\
\textbf{\methodPPRoverE} (Ours)            & \textbf{0.837}  & \textbf{0.887}  & \textbf{0.897}  & \textbf{0.879}  & \underline{0.856} & \textbf{0.871}       \\ \bottomrule
\end{tabular}%
}
\vspace{-10pt} 
\end{wraptable}

\textbf{Single-Model Statistical Criteria Outperform Dual-Model and Resampling Baselines} As shown in \Cref{tab:main_results_by_dataset,tab:main_results_by_model}, entropy attains the highest overall AUROC of 0.812 among all baselines, while Likelihood, LogRank and LastDE maintain robust stability across diverse source models and datasets. 
In contrast, dual-model and resampling-based detectors exhibit substantially inferior performance and severe instability across source models.
The dramatic degradation from LastDE to LastDE++, which normalizes the LastDE score using resampled continuations, isolates the adverse effect of resampling and corroborates our \emph{prompt conditioning gap} analysis in \Cref{sec:resampling-fail}.

\textbf{\methodPPR and Entropy are Complementary Signals} Combining \methodPPR and entropy achieves the best performance. \methodPPRminus already surpasses every baseline, including Entropy itself, confirming that fluctuation magnitude and sequence-level entropy are complementary signals.
Normalizing the \methodPPR by the sequence-level entropy, as in \methodPPRoverE, yields a further improvement over \methodPPRminus on every source model and every dataset except PubMedQA. The outstanding performance of \methodPPRoverE demonstrates that evaluating entropy fluctuations relative to base entropy provides a highly discriminative signal for identifying instruction-conditioned machine text.

\subsection{Robustness Analysis} 
\label{sec:usability_analysis}

\begin{figure}[t]
\centering
\includegraphics[width=0.6\textwidth]{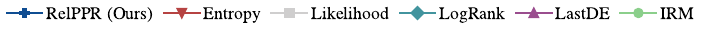}\\[-3.8pt]
\begin{subfigure}[t]{0.33\textwidth}
  \centering
  \includegraphics[width=\textwidth]{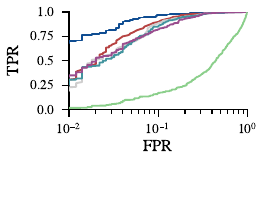}
  \captionsetup{margin={19pt,0pt},skip=-12pt}
  \caption{ROC curve (ELI5)}
  \vspace{-1em}
  \label{fig:auroc_roc_olmo32b_eli5}
\end{subfigure}%
\hfill
\begin{subfigure}[t]{0.315\textwidth}
  \centering
  \includegraphics[width=\textwidth]{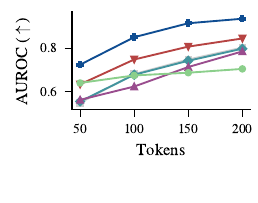}
  \captionsetup{margin={21pt,0pt},skip=-12pt}
  \caption{Sequence Length}
  \vspace{-1em}
  \label{fig:auroc_length_ablation_olmo32b}
\end{subfigure}%
\hfill
\begin{subfigure}[t]{0.33\textwidth}
  \centering
  \includegraphics[width=\textwidth]{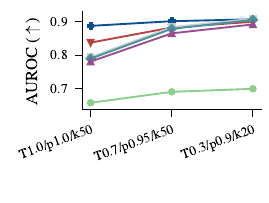}
  \captionsetup{margin={21pt,0pt},skip=-12pt}
  \caption{Decoding Strategy}
  \vspace{-1em}
  \label{fig:auroc_decoding_params_olmo32b}
\end{subfigure}
\caption{Detection Performance across decision thresholds, sequence length and decoding configurations on OLMo-3.1-32B-Instruct. Across all evaluations, \methodPPRoverE consistently outperforms baselines and exhibits notable robustness.
\vspace{-1em}
}
\label{fig:auroc_length_decoding_row}
\end{figure}

Beyond overall detection accuracy, we examine whether \methodPPRoverE remains reliable under conditions that matter for real-world deployment, including skewed decision thresholds, varying text length and decoding configurations, adversarial paraphrasing and evaluator choices.


\textbf{Robustness across Decision Thresholds, Length and Decoding Configurations} 
Beyond overall AUROC, practical deployment also depends on how much recall (true positive rate, TPR) a detector retains at low false positive rate (FPR). \Cref{fig:auroc_roc_olmo32b_eli5} plots the ROC curve (log-scale FPR) on dataset ELI5 for \methodPPRoverE and the top 5 baselines. \methodPPRoverE dominates the low-FPR regime, retaining substantially higher TPR than all baselines as FPR approaches $0.01$.
We further vary generation length in \Cref{fig:auroc_length_ablation_olmo32b} and decoding determinism in \Cref{fig:auroc_decoding_params_olmo32b}. Shorter texts and more stochastic decoding are uniformly harder for all evaluated detectors. Across these settings, \methodPPRoverE leads every baseline at every length tier and every decoding setting. The detailed setup is provided in \Cref{sec:appendix-length-ablation-setup}.

\setlength{\intextsep}{0pt}
\begin{wrapfigure}[]{r}{0.5\textwidth}
\vspace{-5pt}
\centering
\includegraphics[width=0.9\linewidth]{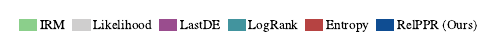}\\[-5pt]
\includegraphics[height=1.1in]{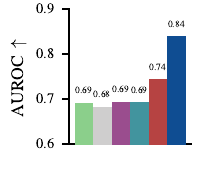}%
\hfill
\includegraphics[height=1.1in]{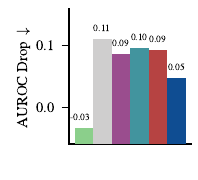}
\vspace{-18pt} 
\caption{ Post-attack AUROC (Left) and AUROC drop (Right) under paraphrasing attack, on OLMo-3.1-32B-Instruct.}
\label{fig:paraphrase_auroc_olmo32b}
\vspace{-8pt}
\end{wrapfigure}

\textbf{Robustness under Paraphrasing Attack} A practical detector must withstand adversarial rewriting. We paraphrase machine text with a T5-based model~\citep{alisetti2021paraphrase}, keeping human text untouched. As shown in \Cref{fig:paraphrase_auroc_olmo32b}, \methodPPRoverE attains the highest post-attack AUROC and lowest drop among declining criteria. This indicates that \methodPPRoverE survives rewriting far better than other single-model statistics. Although IRM's AUROC rises under paraphrasing, it still trails \methodPPRoverE by a wide margin.

\textbf{Robustness Across Evaluator Size and Family, and White-Box Detection}
As shown in \Cref{tab:evaluator_size_ablation}, \methodPPRoverE generalizes well across evaluator architectures and scales, ranking first on seven of nine evaluators. Furthermore, cross-family shifts affect performance more than scaling within a family. We report the results under the white-box setting in \Cref{tab:appendix-whitebox-results-by-dataset,tab:appendix-whitebox-results-by-model}, where dual-model detectors achieve the best performance, as access to the source model restores cross-model discrepancy, while resampling-based methods remain limited by the prompt conditioning gap. Full experimental details and discussion are provided in \Cref{sec:appendix-whitebox-results}.


\section{Conclusion}
We study zero-shot detection of LLM-generated text in the black-box, instruction-conditioned setting, where state-of-the-art dual-model and resampling-based detectors break down.
We trace this failure to two mechanisms.
For dual-model detectors, instruction conditioning shrinks the cross-model discrepancy these detectors rely on, through lower predictive uncertainty or higher cross-model consensus.
For resampling-based detectors, the prompt conditioning gap drives the resampled text away from the candidate.
Motivated by the contrast between the uniform information flow of human writing and the sharp local fluctuations of machine text, we propose \methodPPRoverE, which measures the prominence of local peaks along the entropy trajectory relative to the sequence-level mean entropy.
\methodPPRoverE is parameter-free and resampling-free, requiring only a single evaluator model and one forward pass.
Across seven datasets and five source models, \methodPPRoverE achieves state-of-the-art black-box detection performance and remains robust across decision thresholds, sequence lengths, decoding configurations, paraphrasing attacks, and evaluator models.
\newpage
\section*{Acknowledgments}
This research was supported in part by the Defense Advanced Research Projects Agency (DARPA) under Agreement No. HR0011269E227 and by the University of Virginia Strategic Investment Fund through the Data Analytics Center (SIF192).

\section*{AI Use Statement}
In this work, we used generative AI tools to polish writing. We have not used generative AI tools to generate synthetic 
datasets, help develop theoretical models or conceptual frameworks, propose 
or refine hypotheses, formulate mathematical claims, provide critical 
ingredients for proving mathematical claims, assist in the writing of 
proofs, design or provide feedback on research methodology or experiments, 
implement the proposed \methodPPRoverE{} method, assist with translation, 
clean and reformat the dataset, or support qualitative or thematic data 
analysis. The machine-generated texts in our experiments are produced by the source models listed in \Cref{sec:appendix-detection-regimes} as the object of study, \textbf{not} as AI assistance to the research. We have reviewed all AI-assisted work. The authors carefully revised all AI-polished or AI-drafted text to ensure it accurately reflects our methods and findings; the authors checked all figures and processed data against the raw experimental outputs; and the authors reviewed and tested all AI-assisted code before use. We take responsibility for the final content of this work, including text, claims, or artifacts produced with the aid of generative AI.

\section*{Ethics Statement}
This work studies zero-shot detection of LLM-generated text; we discuss its main ethical considerations below.

\textbf{Data and human subjects}
All human-written text is drawn from publicly available benchmark datasets (XSum, CNN/DailyMail, WikiText-103, SQuAD, WritingPrompts, ELI5, and PubMedQA) that are widely used in prior detection work, and we use them for research purposes in accordance with their respective licenses. WritingPrompts and ELI5 consist of user-generated content collected from Reddit; we use the text only as released, collect no additional personal information, and make no attempt to identify its authors. Machine-generated text was produced by publicly available open-weight models. The study involves no human participants or annotators.

\textbf{Risk of false positives}
Errors of a text detector fall on people: a human author whose writing is flagged as machine-generated may face unwarranted accusations. Our results indicate that this risk is not uniform across domains. On PubMedQA, human text dense in specialized terminology exhibits entropy fluctuations resembling those of machine text (\Cref{sec:main_results}), suggesting elevated false-positive rates on technical writing. Prior work has also shown that detectors can be biased against non-native English writers~\citep{liang2023gpt}; our evaluation is limited to English and does not assess this dimension. We therefore caution that \methodPPRoverE, like other detectors, should not be used as the sole basis for consequential decisions (e.g., academic misconduct findings or content removal), and that its scores should be treated as probabilistic evidence subject to human review.


\bibliography{iclr2027_conference}
\bibliographystyle{iclr2027_conference}

\clearpage

\appendix
\crefalias{section}{appendix} 
\crefalias{subsection}{appendix}
\crefalias{subsubsection}{appendix}
\section{Failure Analysis}\label{sec:appendix-failure-analysis}
\begin{table}[!t]
\centering
\small
\setlength{\tabcolsep}{3.2pt}
\caption[Per-dataset ROC AUC comparison between $\mathrm{Score}_{\mathrm{F}}(s)$ and $N(s)$.]{Per-dataset ROC AUC comparison between $\mathrm{Score}_{\mathrm{F}}(s)$ and $N(s)$ for OLMo-3-32B-generated text. $\Delta = \mathrm{Avg}_{N} - \mathrm{Avg}_{\mathrm{Score}_{\mathrm{F}}}$. Removing the variance denominator produces virtually no change in AUROC ($|\Delta| \le 0.007$).}
\label{tab:fast_detectgpt_numerator_ablation}
\vspace{4pt}
\resizebox{\textwidth}{!}{%
\begin{tabular}{lcccccccccc}
\toprule
\textbf{Condition} & \textbf{Variant} & \textbf{XSum} & \textbf{SQuAD} & \textbf{WritingP} & \textbf{CNN/DM} & \textbf{ELI5} & \textbf{WikiText} & \textbf{PubMedQA} & \textbf{Average} & \textbf{$\Delta$} \\
\midrule
\multirow{2}{*}{w/o Instruction}
  & $\mathrm{Score}_{\mathrm{F}}(s)$        & 0.766 & 0.753 & 0.769 & 0.708 & 0.683 & 0.609 & 0.719 & 0.715 & \multirow{2}{*}{$+$0.007} \\
& $N(s)$   & 0.794 & 0.749 & 0.774 & 0.707 & 0.687 & 0.615 & 0.728 & 0.722 & \\
\cmidrule{1-11}
\multirow{2}{*}{w/ Instruction}
  & $\mathrm{Score}_{\mathrm{F}}(s)$        & 0.569 & 0.514 & 0.494 & 0.497 & 0.556 & 0.509 & 0.465 & 0.515 & \multirow{2}{*}{$0.000$} \\
& $N(s)$   & 0.569 & 0.502 & 0.501 & 0.500 & 0.552 & 0.510 & 0.469 & 0.515 & \\
\bottomrule
\end{tabular}%
}
\end{table}
\subsection{Fast-DetectGPT}\label{sec:appendix-fast-detectgpt}
In the general dual-model formulation where $p_1$ serves as the sampling model and $p_2$ as the scoring model, Fast-DetectGPT~\citep{fastdetectgpt} considers drawing candidate counterfactual tokens $\tilde{x} \sim p_1(\cdot \mid x_{<i})$ at each position $i$. It computes their conditional expectation $\tilde{\mu}_i$ and variance $\tilde{\sigma}_i^2$ under $p_2$:
\begin{equation*}
\tilde\mu_i = \E_{\tilde x \sim p_1(\cdot\mid x_{<i})}\big[\log p_2(\tilde x\mid x_{<i})\big], \qquad
\tilde\sigma_i^2 = \mathrm{Var}_{\tilde x \sim p_1(\cdot\mid x_{<i})}\big[\log p_2(\tilde x\mid x_{<i})\big].
\end{equation*}
The detection score is then defined as the normalized sum of conditional likelihood deviations:
\begin{equation*}
\mathrm{Score}_{\mathrm{F}}(s) = \frac{N(s)}{\sqrt{\sum_{i=1}^L \tilde\sigma_i^2}}, \qquad N(s) = \sum_{i=1}^L \Big[\log p_2(x_i\mid x_{<i}) - \tilde\mu_i\Big],
\end{equation*}
where the numerator $N(s)$ accumulates conditional likelihood deviations, and the denominator normalizes them by the conditional variance. By definition, $\tilde\mu_i$ is the negative of the per-position cross-entropy, so summing over $i$ gives:
\begin{equation*}
\sum_{i=1}^L \tilde\mu_i = -L\cdot H(p_1, p_2; s), \qquad \sum_{i=1}^L \log p_2(x_i\mid x_{<i}) = -L\cdot I_{p_2}(s).
\end{equation*}
Substituting both into $N(s)$ yields:
\begin{equation*}
N(s) =  L\cdot \big(H(p_1, p_2; s) - I_{p_2}(s)\big).
\end{equation*}

\Cref{tab:fast_detectgpt_numerator_ablation} compares $\mathrm{Score}_{\mathrm{F}}(s)$ against $N(s)$ across generation conditions and datasets, to test whether the variance normalization denominator drives the performance collapse on instruction-tuned generations. As shown in the table, removing the denominator produces virtually identical detection performance across all settings, validating our analytical focus on decomposing $N(s)$ in \Cref{sec:fast-detectgpt}.

\subsection{Binoculars and DNA-DetectLLM}\label{sec:appendix-binoculars-dna-detectllm}
Rather than measuring a difference, Binoculars~\citep{binoculars} and DNA-DetectLLM~\citep{dna-detectllm} employ a ratio-based formulation. Following their official code implementations, Binoculars defines its score as the negative ratio of log-perplexity to log cross-perplexity given model $p_1$ and model $p_2$,:
\begin{equation*}
\mathrm{Score}_{\mathrm{B}}(s) = -\frac{\log \mathrm{PPL}_{p_2}(s)}{\log \mathrm{X}\text{-}\mathrm{PPL}_{p_1, p_2}(s)} = -\frac{I_{p_2}(s)}{H(p_1, p_2; s)}.
\end{equation*}
DNA-DetectLLM additionally repairs $s$ into a greedily re-decoded sequence $\tilde{s}$ under the scoring model, $\tilde{s}_t = \arg\max_{v \in \mathcal{V}} p_2(v \mid s_{<t})$, and defines the conditional surprisal of $\tilde{s}$ given $s$ as the surprisal of the repaired sequence evaluated position-by-position against the original prefixes,
\begin{equation*}
I_{p_2}(\tilde{s} \mid s) = -\frac{1}{L}\sum_{i=1}^L \log p_2(\tilde{x}_i \mid x_{<i}).
\end{equation*}
The score is defined as:
\begin{equation*}
\mathrm{Score}_{\mathrm{D}}(s) = -\frac{\log \mathrm{PPL}_{p_2}(s) + \log \mathrm{PPL}_{p_2}(\tilde{s} \mid s)}{2 \cdot \log \mathrm{X}\text{-}\mathrm{PPL}_{p_1, p_2}(s)} = -\frac{I_{p_2}(s) + I_{p_2}(\tilde{s} \mid s)}{2 \cdot H(p_1, p_2; s)}.
\end{equation*}

\subsection{Cross-Entropy Decomposition}\label{sec:appendix-cross-entropy-decomposition}
At each position $i$, the cross-entropy, entropy, and KL divergence between the sampling distribution $p_1(\cdot \mid x_{<i})$ and the scoring distribution $p_2(\cdot \mid x_{<i})$ satisfy the standard identity
\begin{equation*}
H(p_1, p_2)_i = H(p_1)_i + D_{\mathrm{KL}}(p_1, p_2)_i,
\end{equation*}
which holds for any pair of distributions over the same support $\mathcal{V}$. Averaging both sides over positions $i = 1, \dots, L$ preserves the identity by linearity:
\begin{equation*}
H(p_1, p_2; s) = \frac{1}{L}\sum_{i=1}^L H(p_1)_i + \frac{1}{L}\sum_{i=1}^L D_{\mathrm{KL}}(p_1, p_2)_i = H(p_1; s) + D_{\mathrm{KL}}(p_1, p_2; s),
\end{equation*}
where $D_{\mathrm{KL}}(p_1, p_2; s) = \frac{1}{L}\sum_{i=1}^L D_{\mathrm{KL}}\big(p_1(\cdot\mid x_{<i}) \parallel p_2(\cdot\mid x_{<i})\big)$ denotes the sequence-level KL divergence.

\subsection{Detection Score and Statistic Distributions on Additional Datasets}\label{sec:appendix-fdgpt-components-other-datasets}
\begin{wrapfigure}[11]{r}{0.5\textwidth}
\centering
\includegraphics[width=0.48\textwidth]{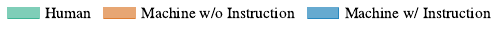}\\[-4pt]
\begin{subfigure}[t]{0.23\textwidth}
  \centering
  \includegraphics[width=\linewidth]{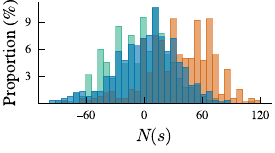}
  \caption{Fast-DetectGPT}
\end{subfigure}%
\hfill%
\begin{subfigure}[t]{0.23\textwidth}
  \centering
  \includegraphics[width=\linewidth]{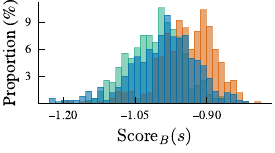}
  \caption{Binoculars}
\end{subfigure}
\caption{Distributions of the Fast-DetectGPT numerator $N(s)$ and Binoculars score $\mathrm{Score}_{\mathrm{B}}(s)$ on XSum (OLMo-3.1-32B-Instruct).}
\label{fig:dual-model-baselines-xsum}
\end{wrapfigure}
\begin{figure}[!tbp]
\centering
\includegraphics[width=2.0in]{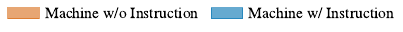}\\[-3pt]
\begin{subfigure}[t]{1.18in}
  \centering
  \includegraphics[width=1.18in]{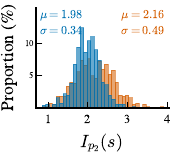}
  \captionsetup{margin={19pt,0pt}}
\end{subfigure}%
\hspace{2pt}%
\begin{subfigure}[t]{1.05in}
  \centering
  \includegraphics[width=1.05in]{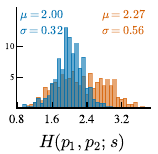}
  \captionsetup{skip=4.5pt}
\end{subfigure}%
\hspace{2pt}%
\begin{subfigure}[t]{1.05in}
  \centering
  \includegraphics[width=1.05in]{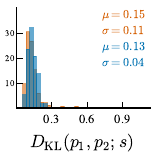}
  \captionsetup{skip=4.5pt}
\end{subfigure}%
\hspace{2pt}%
\begin{subfigure}[t]{1.05in}
  \centering
  \includegraphics[width=1.05in]{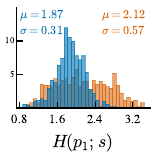}
  \captionsetup{skip=4.5pt}
\end{subfigure}%
\hspace{2pt}%
\begin{subfigure}[t]{1.05in}
  \centering
  \includegraphics[width=1.05in]{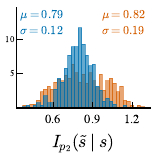}
  \captionsetup{skip=4.5pt}
\end{subfigure}
\caption{%
  Distributions of the five information-theoretic components from Section~\ref{sec:dual-model-detectors} on ELI5, for machine text generated by OLMo-3.1-32B-Instruct without and with the instruction prompt. The y-axis shows the percentage of samples per bin. $\mu$ and $\sigma$ denote the mean and standard deviation.
}
\label{fig:fdgpt-components-eli5}
\end{figure}
\begin{figure}[!tbp]
\centering
\includegraphics[width=2.0in]{figure/fdgpt_components_legend.pdf}\\[-3pt]
\begin{subfigure}[t]{1.18in}
  \centering
  \includegraphics[width=1.18in]{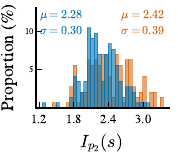}
  \captionsetup{margin={19pt,0pt}}
\end{subfigure}%
\hspace{2pt}%
\begin{subfigure}[t]{1.05in}
  \centering
  \includegraphics[width=1.05in]{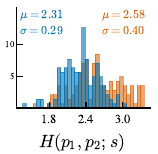}
  \captionsetup{skip=4.5pt}
\end{subfigure}%
\hspace{2pt}%
\begin{subfigure}[t]{1.05in}
  \centering
  \includegraphics[width=1.05in]{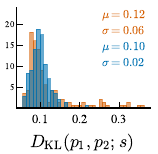}
  \captionsetup{skip=4.5pt}
\end{subfigure}%
\hspace{2pt}%
\begin{subfigure}[t]{1.05in}
  \centering
  \includegraphics[width=1.05in]{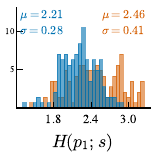}
  \captionsetup{skip=4.5pt}
\end{subfigure}%
\hspace{2pt}%
\begin{subfigure}[t]{1.05in}
  \centering
  \includegraphics[width=1.05in]{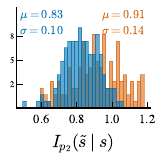}
  \captionsetup{skip=4.5pt}
\end{subfigure}
\caption{%
  Distributions of the five information-theoretic components from Section~\ref{sec:dual-model-detectors} on SQuAD, for machine text generated by OLMo-3.1-32B-Instruct without and with  the instruction prompt. The y-axis shows the percentage of samples per bin. $\mu$ and $\sigma$ denote the mean and standard deviation.
}
\label{fig:fdgpt-components-squad}
\end{figure}
\begin{figure}[!tbp]
\centering
\includegraphics[width=2.0in]{figure/fdgpt_components_legend.pdf}\\[-3pt]
\begin{subfigure}[t]{1.18in}
  \centering
  \includegraphics[width=1.18in]{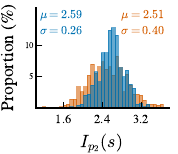}
  \captionsetup{margin={19pt,0pt},skip=4.5pt}
\end{subfigure}%
\hspace{2pt}%
\begin{subfigure}[t]{1.05in}
  \centering
  \includegraphics[width=1.05in]{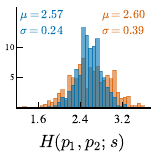}
  \captionsetup{skip=4.5pt}
\end{subfigure}%
\hspace{2pt}%
\begin{subfigure}[t]{1.05in}
  \centering
  \includegraphics[width=1.05in]{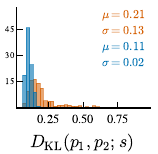}
  \captionsetup{skip=4.5pt}
\end{subfigure}%
\hspace{2pt}%
\begin{subfigure}[t]{1.05in}
  \centering
  \includegraphics[width=1.05in]{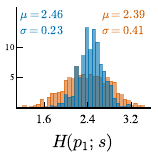}
  \captionsetup{skip=4.5pt}
\end{subfigure}%
\hspace{2pt}%
\begin{subfigure}[t]{1.05in}
  \centering
  \includegraphics[width=1.05in]{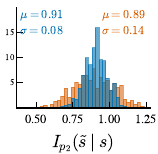}
  \captionsetup{skip=4.5pt}
\end{subfigure}
\caption{%
  Distributions of the five information-theoretic components from Section~\ref{sec:dual-model-detectors} on CNN/DailyMail, for machine text generated by OLMo-3.1-32B-Instruct without (orange) and with (blue) the instruction prompt. The y-axis shows the percentage of samples per bin. $\mu$ and $\sigma$ denote the mean and standard deviation.
}
\label{fig:fdgpt-components-cnn_dailymail}
\end{figure}
\begin{figure}[!tbp]
\centering
\includegraphics[width=2.0in]{figure/fdgpt_components_legend.pdf}\\[-3pt]
\begin{subfigure}[t]{1.18in}
  \centering
  \includegraphics[width=1.18in]{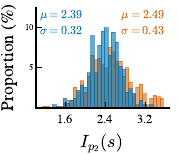}
  \captionsetup{margin={19pt,0pt}}
\end{subfigure}%
\hspace{2pt}%
\begin{subfigure}[t]{1.05in}
  \centering
  \includegraphics[width=1.05in]{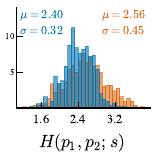}
  \captionsetup{skip=4.5pt}
\end{subfigure}%
\hspace{2pt}%
\begin{subfigure}[t]{1.05in}
  \centering
  \includegraphics[width=1.05in]{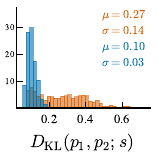}
  \captionsetup{skip=4.5pt}
\end{subfigure}%
\hspace{2pt}%
\begin{subfigure}[t]{1.05in}
  \centering
  \includegraphics[width=1.05in]{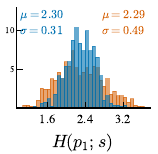}
  \captionsetup{skip=4.5pt}
\end{subfigure}%
\hspace{2pt}%
\begin{subfigure}[t]{1.05in}
  \centering
  \includegraphics[width=1.05in]{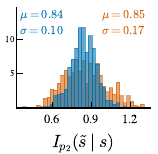}
  \captionsetup{skip=4.5pt}
\end{subfigure}
\caption{%
  Distributions of the five information-theoretic components from Section~\ref{sec:dual-model-detectors} on WikiText, for machine text generated by OLMo-3.1-32B-Instruct without (orange) and with (blue) the instruction prompt. The y-axis shows the percentage of samples per bin. $\mu$ and $\sigma$ denote the mean and standard deviation.
}
\label{fig:fdgpt-components-wikitext}
\end{figure}
\begin{figure}[!tbp]
\centering
\includegraphics[width=2.0in]{figure/fdgpt_components_legend.pdf}\\[-3pt]
\begin{subfigure}[t]{1.18in}
  \centering
  \includegraphics[width=1.18in]{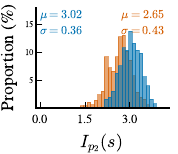}
  \captionsetup{margin={19pt,0pt}}
\end{subfigure}%
\hspace{2pt}%
\begin{subfigure}[t]{1.05in}
  \centering
  \includegraphics[width=1.05in]{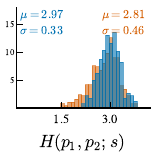}
  \captionsetup{skip=4.5pt}
\end{subfigure}%
\hspace{2pt}%
\begin{subfigure}[t]{1.05in}
  \centering
  \includegraphics[width=1.05in]{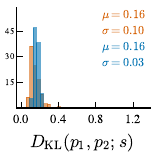}
  \captionsetup{skip=4.5pt}
\end{subfigure}%
\hspace{2pt}%
\begin{subfigure}[t]{1.05in}
  \centering
  \includegraphics[width=1.05in]{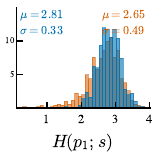}
  \captionsetup{skip=4.5pt}
\end{subfigure}%
\hspace{2pt}%
\begin{subfigure}[t]{1.05in}
  \centering
  \includegraphics[width=1.05in]{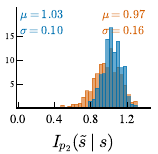}
  \captionsetup{skip=4.5pt}
\end{subfigure}
\caption{%
  Distributions of the five information-theoretic components from Section~\ref{sec:dual-model-detectors} on WritingPrompts, for machine text generated by OLMo-3.1-32B-Instruct without (orange) and with (blue) the instruction prompt. The y-axis shows the percentage of samples per bin. $\mu$ and $\sigma$ denote the mean and standard deviation.
}
\label{fig:fdgpt-components-writingprompts}
\end{figure}

\Cref{fig:dual-model-baselines-xsum} shows the distributions of the Fast-DetectGPT numerator $N(s)$ and the Binoculars score $\mathrm{Score}_{\mathrm{B}}(s)$. Under instruction conditioning, the machine-text distributions shift markedly toward lower values and largely overlap with the human-text distributions, which severely reduces detection separability. We further report the distributions of the five statistics introduced in \Cref{sec:dual-model-detectors} on the remaining five datasets in \Cref{fig:fdgpt-components-eli5,fig:fdgpt-components-cnn_dailymail,fig:fdgpt-components-squad,fig:fdgpt-components-wikitext,fig:fdgpt-components-writingprompts}.
Instruction conditioning consistently lowers the cross-entropy $H(p_1, p_2; s)$.
At a finer granularity, this shrinkage is driven predominantly by a drop in the KL divergence $D_{\mathrm{KL}}(p_1, p_2; s)$ on CNN/DailyMail and WikiText, and primarily by a reduction in the model entropy $H(p_1; s)$ on SQuAD and ELI5.
The only exception is WritingPrompts, where the detection scores instead decrease because the surprisal $I_{p_2}(s)$ increases, as the invisible prompt induces more diverse stories.

\subsection{\methodPPRoverE Distributions on Additional Datasets}\label{sec:appendix-ppr-other-datasets}
\begin{figure}[!tbp]
\centering
\includegraphics[width=2.1in]{figure/relppr_legend.pdf}\\[-2pt]
\includegraphics[height=0.95in]{figure/relppr_ylabel.pdf}%
\begin{subfigure}[t]{1.32in}
  \centering
  \includegraphics[width=1.32in]{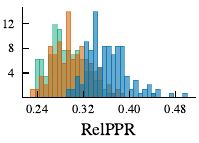}
  \captionsetup{margin={10.5pt,-0.5pt},skip=4.5pt}
  \caption{SQuAD}
\end{subfigure}%
\hfill%
\begin{subfigure}[t]{1.32in}
  \centering
  \includegraphics[width=1.32in]{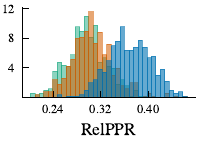}
  \captionsetup{margin={10.5pt,-0.5pt},skip=4.5pt}
  \caption{WikiText}
\end{subfigure}%
\hfill%
\begin{subfigure}[t]{1.32in}
  \centering
  \includegraphics[width=1.32in]{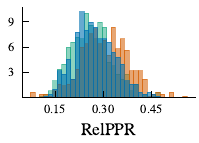}
  \captionsetup{margin={10.5pt,-0.5pt},skip=4.5pt}
  \caption{PubMedQA}
\end{subfigure}
\caption{%
  Distributions of \methodPPRoverE across SQuAD, WikiText, and PubMedQA for human text alongside machine-generated text from OLMo-3.1-32B-Instruct.
}
\label{fig:ppr-appendix}
\end{figure}

\Cref{fig:ppr-appendix} displays the corresponding distributions for the remaining three benchmark datasets. On SQuAD and WikiText, \methodPPRoverE clearly separates human writing from machine-generated text and exhibits the characteristic shift toward higher values under instruction conditioning.

\section{Experimental Setup and Implementation Details}\label{sec:appendix-implementation-details}\label{sec:appendix-prompt-templates}

In this section, we present the complete experimental setup, including prompt templates, text generation protocols, length-control mechanisms, and evaluator model pairings used across our experiments.

\subsection{Detection Settings and Evaluator Models}\label{sec:appendix-detection-regimes}
\textbf{Source Models.}
We use five instruction-tuned LLMs as source models: OLMo-3-7B-Instruct, OLMo-3.1-32B-Instruct~\citep{olmo2025olmo}, Qwen3.5-35B-A3B~\citep{qwen3.5}, Gemma-4-31B-it~\citep{team2026gemma}, Mistral-Small-4-119B~\citep{mistral-small-4}.

\textbf{Black-Box Setting.}
In the black-box setting, the detector operates without knowledge of the source model. Single-model statistical criteria employ Qwen3.5-2B-Base as the evaluator (and CoEdit-Large for GECScore). Dual-model baselines use Qwen3.5-2B-Base and Qwen3.5-2B-Instruct as $p_1$ and $p_2$, respectively.  To assess sensitivity to
the evaluator choice, we further vary the evaluator across sizes and families,
including Qwen3.5 (2B, 9B, 35B-MoE), Gemma~4 (12B, 31B, 26B-MoE), Ministral~3
(8B), Falcon3 (7B), and Falcon (7B), with results reported in \Cref{tab:evaluator_size_ablation}.

\textbf{White-Box Setting.}
In the white-box regime, the detector directly utilizes the generative source model ($M = M_{\mathrm{source}}$). For dual-model detectors, the source instruction-tuned model serves as $p_2$, while its same-lineage foundation base counterpart at the same parameter scale serves as the reference/observer model $p_1$ (e.g., OLMo-3-7B-Base for OLMo-3-7B-Instruct, Qwen3.5-35B-A3B-Base for Qwen3.5-35B-A3B). Mistral-Small-4-119B is excluded from the white-box setting, as no same-lineage base checkpoint is available for it; GECScore is likewise excluded, since it always scores against a fixed off-the-shelf grammar-correction model and therefore has no white-box variant.

\subsection{Prompt Templates and Decoding Configuration}
\Cref{tab:appendix-prompt-templates} details the instruction prompt template for each benchmark dataset. In our main experiments, all text generations are carried out using the vLLM engine with $T = 1.0$, $p = 1.0$, and top-$k = 50$.

\subsection{Text Length Control and Symmetrical Trimming}\label{sec:appendix-length-control}
Systematic length discrepancies between human and machine text can introduce trivial shortcut features into detection. To ensure rigorous, length-unbiased evaluation, we enforce a multi-stage length-control protocol:
\begin{enumerate}
    \item \emph{Document pre-filtering:} Human documents shorter than a minimum word threshold (200 words for standard datasets, and $150$ words for ELI5) are filtered out to guarantee sufficient text length (PubMedQA is not filtered by length, since the sequence in it is short).
    \item \emph{Generation constraint:} The prompt parameter \texttt{\{word\_target\}} is uniformly set to $150$ words across six datasets, and to $30$ words for PubMedQA to match its considerably shorter human answers. During generation, we set \texttt{min\_tokens} to 150 for datasets except PubMedQA, and to 30 for PubMedQA, alongside an upper bound of \texttt{max\_new\_tokens} $= 200$ for all datasets.
    \item \emph{Post-trimming:} Following \citet{fastdetectgpt}, after generating machine continuations, both the human completion and machine text are encoded into token sequences under the generative model's tokenizer and trimmed down to their shared minimum token length: $L = \min(|s_{\mathrm{human}}|, |s_{\mathrm{machine}}|)$. Both trimmed sequences are then decoded back to text. 
\end{enumerate}
This guarantees that for every candidate pair, the human and machine texts have the \emph{exact same sequence length} under the evaluation tokenizer. We draw $500$ paired samples ($1{,}000$ texts total) per dataset and model, except for SQuAD where $143$ pairs are yielded due to pre-filtering length constraint. The resulting sample lengths for XSum, WritingPrompts, SQuAD, and CNN/DailyMail are uniformly 200 tokens, with detailed token length distributions on ELI5 and PubMedQA provided in \Cref{tab:appendix-eli5-token-lengths}.

\begin{table}[t]
\centering
\small
\setlength{\tabcolsep}{10pt}
\caption{Token length distributions (Mean $\pm$ Std) on the ELI5 and PubMedQA across all evaluated instruction models ($n = 500$ pairs per model), measured with each source model's own tokenizer.}
\label{tab:appendix-eli5-token-lengths}
\resizebox{\textwidth}{!}{
\begin{tabular}{lccccc}
\toprule
 & \textbf{OLMo-7B} & \textbf{OLMo-32B} & \textbf{Gemma-4-31B} & \textbf{Qwen-35B-MoE} & \textbf{Mistral-119B} \\
\midrule
ELI5 & $178.9 \pm 12.8$ & $174.4 \pm 13.6$ & $171.2 \pm 9.7$ & $170.2 \pm 11.8$ & $182.8 \pm 14.6$ \\
PubMedQA & $33.3 \pm 6.1$ & $34.3 \pm 8.9$ & $32.0 \pm 4.7$ & $32.9 \pm 6.1$ & $32.8 \pm 6.2$ \\
\bottomrule
\end{tabular}
}
\end{table}
\subsection{Length-Ablation and Decoding-Ablation Experimental Setup}\label{sec:appendix-length-ablation-setup}
\begin{table}[t]
\centering
\small
\setlength{\tabcolsep}{6pt}
\caption{Instruction prompt templates across all seven benchmark datasets. \texttt{\{prefix\}} represents the conditioning context, and \texttt{\{word\_target\}} denotes the expected word count (150 words, except PubMedQA at 30 words).}
\label{tab:appendix-prompt-templates}
\vspace{4pt}
\resizebox{\textwidth}{!}{%
\begin{tabular}{lp{12.5cm}}
\toprule
\textbf{Dataset} & \textbf{Instruction Prompt Template} \\
\midrule
XSum & ``\textit{Continue the following text for about} \texttt{\{word\_target\}} \textit{words. Output ONLY the continuation -- do not repeat or restate any of the given text:} \texttt{\{prefix\}}'' \\
\addlinespace[3pt]
CNN/DM & ``\textit{Continue the following text for about} \texttt{\{word\_target\}} \textit{words. Output ONLY the continuation -- do not repeat or restate any of the given text:} \texttt{\{prefix\}}'' \\
\addlinespace[3pt]
WikiText & ``\textit{Continue the following text for about} \texttt{\{word\_target\}} \textit{words. Output ONLY the continuation -- do not repeat or restate any of the given text:} \texttt{\{prefix\}}'' \\
\addlinespace[3pt]
SQuAD & ``\textit{Continue the following text for about} \texttt{\{word\_target\}} \textit{words. Output ONLY the continuation -- do not repeat or restate any of the given text:} \texttt{\{prefix\}}'' \\
\addlinespace[3pt]
WritingPrompts & ``\textit{Please write a story of about} \texttt{\{word\_target\}} \textit{words based on the following writing prompt. Output ONLY the story -- do not repeat or restate any of the given text:} \texttt{\{prefix\}}'' \\
\addlinespace[3pt]
ELI5 & ``\textit{Please explain the following question in about} \texttt{\{word\_target\}} \textit{words. Output ONLY the explanation -- do not repeat or restate any of the given text:} \texttt{\{prefix\}}'' \\
\addlinespace[3pt]
PubMedQA & ``\textit{Please answer the question in about} \texttt{\{word\_target\}} \textit{words. Output ONLY the answer -- do not repeat or restate any of the given text:} \texttt{\{prefix\}}'' \\
\bottomrule
\end{tabular}%
}
\end{table}
For each tier, we re-generate machine continuations with the prompt's \texttt{\{word\_target\}} set to a length-matched word count (40, 80, 120 words), set \texttt{min\_tokens} to this same value, and cap \texttt{max\_new\_tokens} at the token tier itself, forcing a continuation of approximately the target length. Human and machine texts are then symmetrically trimmed as in \Cref{sec:appendix-length-control}. All other settings are unchanged from the main experiments. This procedure is applied in the black-box setting, across the XSum, WritingPrompts, SQuAD, WikiText, and CNN/DailyMail datasets; ELI5 and PubMedQA are excluded since their human documents are shorter.

For the decoding ablation, we re-generate machine continuations under three decoding configurations spanning the sampling-randomness spectrum: $(T{=}1.0, p{=}1.0, k{=}50)$, $(T{=}0.7, p{=}0.95, k{=}50)$, and $(T{=}0.3, p{=}0.9, k{=}20)$, with all other generation settings (length control, datasets, source models) unchanged from the main experiments. Human and machine texts are again symmetrically trimmed as in \Cref{sec:appendix-length-control}, and each baseline is re-scored at every decoding configuration across all seven datasets.

\subsection{Robustness Across Evaluator Size and Family}
\label{sec:appendix-evaluator-robustness}

We assess detection performance across varying evaluator models while 
fixing the source model to OLMo-3.1-32B-Instruct. As shown in Table~\ref{tab:evaluator_size_ablation}, 
\methodPPRoverE demonstrates superior generalizability across diverse evaluator 
architectures and parameter scales, securing the top performance on seven 
of the nine evaluators and ranking second on another. Notably, while 
scaling the evaluator size within the same family yields marginal impact 
on detection performance, cross-family shifts induce substantially more 
pronounced variations. Additionally, within the Qwen3.5 family, 
transitioning from dense architectures to the MoE variant visibly degrades 
the detection efficacy of all baselines.

\begin{table}[t]
\centering
\small
\setlength{\tabcolsep}{4pt}
\caption{Zero-shot detection AUROC on OLMo-3.1-32B-Instruct (black-box, avg. over all 7 datasets) as the evaluator model varies in size and family (Qwen3.5~\citep{qwen3.5}, Gemma 4~\citep{team2026gemma}, Ministral 3~\citep{liu2026ministral}, and Falcon~\citep{almazrouei2023falcon, falcon3}). Best result per column is shown in \textbf{bold}, second-best is \underline{underlined}.
}
\label{tab:evaluator_size_ablation}
\resizebox{\textwidth}{!}{%
\begin{tabular}{lccccccccc c}
\hline
\multirow{2}{*}{\textbf{Methods}} & \textbf{Qwen3.5} & \textbf{Qwen3.5} & \textbf{Qwen3.5} & \textbf{Gemma4} & \textbf{Gemma4} & \textbf{Gemma4} & \textbf{Falcon3} & \textbf{Ministral3} & \textbf{Falcon} & \multirow{2}{*}{\textbf{Avg}} \\
                          & \textbf{2B}       & \textbf{9B}       & \textbf{35B-MoE}  & \textbf{12B}     & \textbf{31B}     & \textbf{26B-MoE} & \textbf{7B}       & \textbf{8B}           & \textbf{7B}      &                      \\ \hline
Entropy & \underline{0.837} & \underline{0.847} & \underline{0.788} & \underline{0.804} & \underline{0.780} & \underline{0.765} & \textbf{0.761} & 0.788 & 0.414 & \underline{0.754} \\
Likelihood & 0.794 & 0.814 & 0.746 & 0.722 & 0.728 & 0.720 & 0.671 & 0.772 & 0.526 & 0.721 \\
LogRank & 0.789 & 0.814 & 0.749 & 0.732 & 0.731 & 0.723 & 0.687 & 0.770 & 0.535 & 0.726 \\
LastDE & 0.781 & 0.812 & 0.755 & 0.761 & 0.769 & 0.760 & 0.684 & \underline{0.796} & 0.657 & 0.753 \\
Fast-DetectGPT & 0.515 & 0.513 & 0.456 & 0.507 & 0.230 & 0.301 & 0.653 & 0.625 & 0.825 & 0.514 \\
Binoculars & 0.521 & 0.526 & 0.472 & 0.683 & 0.652 & 0.523 & 0.649 & 0.642 & \underline{0.827} & 0.611 \\
DNA-DetectLLM & 0.515 & 0.502 & 0.459 & 0.678 & 0.630 & 0.518 & 0.729 & 0.566 & \textbf{0.855} & 0.606 \\
\textbf{\methodPPRoverE} (Ours) & \textbf{0.887} & \textbf{0.863} & \textbf{0.826} & \textbf{0.805} & \textbf{0.804} & \textbf{0.804} & \underline{0.759} & \textbf{0.835} & 0.602 & \textbf{0.798} \\
\hline
\end{tabular}%
}
\end{table}

\begin{table}[t]
\centering
\small
\setlength{\tabcolsep}{5pt}
\caption{Detection performance (AUROC) by dataset (averaged across four source models) under white-box setting. The best and second-best results are shown in \textbf{bold} and \underline{underlined}, respectively.}
\label{tab:appendix-whitebox-results-by-dataset}
\begin{tabular}{@{} l ccccccc c @{}}
\toprule
\textbf{Baseline} & \textbf{XSum} & \textbf{WritingP} & \textbf{SQuAD} & \textbf{WikiText} & \textbf{CNN/DM} & \textbf{ELI5} & \textbf{PubMedQA} & \textbf{Avg} \\
\midrule
Entropy & 0.803 & 0.856 & 0.589 & 0.683 & 0.724 & 0.855 & 0.667 & 0.740 \\
Likelihood & 0.931 & 0.911 & 0.708 & 0.913 & 0.920 & 0.907 & 0.574 & 0.838 \\
LogRank & 0.926 & 0.901 & 0.697 & 0.909 & 0.915 & 0.896 & 0.601 & 0.835 \\
LRR & 0.798 & 0.797 & 0.614 & 0.767 & 0.777 & 0.811 & 0.620 & 0.741 \\
LastDE & 0.887 & 0.882 & 0.794 & 0.941 & 0.908 & 0.890 & 0.534 & 0.834 \\
LastDE++ & 0.416 & 0.555 & 0.245 & 0.414 & 0.485 & 0.381 & 0.426 & 0.418 \\
DNA-GPT & 0.739 & 0.783 & 0.561 & 0.877 & 0.771 & 0.711 & 0.535 & 0.711 \\
Fast-DetectGPT & 0.817 & 0.892 & 0.708 & 0.835 & 0.823 & 0.767 & 0.482 & 0.760 \\
Binoculars & 0.956 & \underline{0.990} & \underline{0.849} & \underline{0.952} & 0.948 & \underline{0.932} & \underline{0.671} & \underline{0.900} \\
DNA-DetectLLM & \underline{0.964} & \textbf{0.992} & 0.845 & 0.946 & \underline{0.949} & 0.915 & 0.584 & 0.885 \\
IRM & \textbf{0.966} & 0.962 & \textbf{0.941} & \textbf{0.974} & \textbf{0.958} & \textbf{0.972} & \textbf{0.679} & \textbf{0.922} \\
\midrule
\methodPPR (Ours) & 0.382 & 0.461 & 0.481 & 0.405 & 0.405 & 0.291 & 0.405 & 0.404 \\
\methodPPRminus (Ours) & 0.825 & 0.893 & 0.608 & 0.700 & 0.737 & 0.887 & 0.648 & 0.757 \\
\methodPPRoverE (Ours) & 0.769 & 0.883 & 0.617 & 0.663 & 0.689 & 0.903 & 0.567 & 0.727 \\
\methodPPRS (Ours) & 0.935 & 0.920 & 0.716 & 0.925 & 0.926 & 0.913 & 0.548 & 0.840 \\
\bottomrule
\end{tabular}
\end{table}

\subsection{White-Box Detection Results}\label{sec:appendix-whitebox-results}
Single-model criteria employ the source model itself as the evaluator, whereas dual-model configurations follow \Cref{sec:appendix-detection-regimes}.
In addition to \methodPPR, \methodPPRminus, and \methodPPRoverE, we test whether \methodPPR can offer complementary signals over standard likelihood by pairing it with sequence-level surprisal, yielding the variant \methodPPRS:
\begin{equation*}
\mathrm{Score}_{\mathrm{\methodPPRS}}(s) = \mathrm{\methodPPR}(s) - I_p(s).
\end{equation*}

\Cref{tab:appendix-whitebox-results-by-dataset,tab:appendix-whitebox-results-by-model} report the detection results under white-box setting.
Unlike the black-box setting, where \methodPPRoverE is state-of-the-art, dual-model detectors achieves the best performance under white-box setting, and IRM attains the best or second-best score in nearly every source model and dataset. 
This is expected: white-box access lets the evaluator model that matches the source model assign very high confidence to the machine text, while the other evaluator model cannot reach a comparable confidence, which restores cross-model discrepancy between the two evaluator models.
Notably, since detectors still cannot access the user prompt under the white-box setting, the prompt conditioning gap described in \Cref{sec:resampling-fail} still persists, so resampling-based baselines remain ineffective under the white-box setting.
While \methodPPRoverE itself is not competitive in this regime, \methodPPRS ranks best among all single-model criteria, confirming that \methodPPR can still serve as a complementary signal under the white-box setting.

\begin{table}[t]
\centering
\small
\setlength{\tabcolsep}{5pt}
\caption{Detection performance (AUROC) by source model (averaged across seven datasets) under white-box setting. The best and second-best results are shown in \textbf{bold} and \underline{underlined}, respectively.}
\label{tab:appendix-whitebox-results-by-model}
\begin{tabular}{@{} l cccc c @{}}
\toprule
\multirow{2}{*}{\textbf{Baseline}} & \textbf{OLMo3} & \textbf{OLMo3.1} & \textbf{Gemma4} & \textbf{Qwen3.5} & \multirow{2}{*}{\textbf{Avg}} \\
                          & \textbf{7B}     & \textbf{32B}      & \textbf{31B}     & \textbf{35B}      &                      \\ \midrule
Entropy & 0.791 & 0.805 & 0.552 & 0.810 & 0.740 \\
Likelihood & 0.901 & 0.870 & 0.776 & 0.803 & 0.838 \\
LogRank & 0.906 & 0.866 & 0.758 & 0.809 & 0.835 \\
LRR & 0.791 & 0.791 & 0.590 & 0.790 & 0.741 \\
LastDE & 0.893 & 0.884 & 0.745 & 0.813 & 0.834 \\
LastDE++ & 0.454 & 0.419 & 0.357 & 0.440 & 0.418 \\
DNA-GPT & 0.808 & 0.765 & 0.711 & 0.560 & 0.711 \\
Fast-DetectGPT & \underline{0.934} & \underline{0.913} & 0.466 & 0.729 & 0.760 \\
Binoculars & \textbf{0.947} & 0.924 & \textbf{0.949} & 0.779 & \underline{0.900} \\
DNA-DetectLLM & \textbf{0.947} & \underline{0.913} & \underline{0.926} & 0.753 & 0.885 \\
IRM & 0.928 & \textbf{0.952} & 0.869 & \textbf{0.938} & \textbf{0.922} \\
\midrule
\methodPPR (Ours) & 0.317 & 0.300 & 0.571 & 0.429 & 0.404 \\
\methodPPRminus (Ours) & 0.774 & 0.801 & 0.625 & 0.826 & 0.757 \\
\methodPPRoverE (Ours) & 0.678 & 0.744 & 0.677 & \underline{0.811} & 0.727 \\
\methodPPRS (Ours) & 0.897 & 0.870 & 0.780 & 0.814 & 0.840 \\
\bottomrule
\end{tabular}
\end{table}

\end{document}

%% file: math_commands.tex
\usepackage{amsmath,amsfonts,bm}

\def\eqref#1{equation~\ref{#1}}

\def\1{\bm{1}}

\DeclareMathAlphabet{\mathsfit}{\encodingdefault}{\sfdefault}{m}{sl}
\SetMathAlphabet{\mathsfit}{bold}{\encodingdefault}{\sfdefault}{bx}{n}

\newcommand{\E}{\mathbb{E}}

